\documentclass[10pt,twocolumn,letterpaper]{article}

\usepackage{cvpr}
\usepackage{times}
\usepackage{epsfig}
\usepackage{graphicx}
\usepackage{amsmath}
\usepackage{amssymb}
\usepackage{subfigure}
\usepackage{multirow}

\usepackage[breaklinks=true,bookmarks=false]{hyperref}

\cvprfinalcopy 

\def\cvprPaperID{****} 

\begin{document}

\title{Trimmed Action Recognition, Dense-Captioning Events in Videos, and Spatio-temporal Action Localization with Focus on ActivityNet Challenge 2019}

\author{Zhaofan Qiu, Dong Li, Yehao Li, Qi Cai, Yingwei Pan, and Ting Yao\\
         JD AI Reseach, Beijing, China\\
         {\tt\small panyw.ustc@gmail.com}
}

\maketitle

\begin{abstract}
This notebook paper presents an overview and comparative analysis of our systems designed for the following three tasks in ActivityNet Challenge 2019: trimmed action recognition, dense-captioning events in videos, and spatio-temporal action localization.

\textbf{Trimmed Action Recognition (Kinetics)}: We investigate and exploit multiple spatio-temporal clues for trimmed action recognition task, i.e., frame, short video clip and motion (optical flow) by leveraging 2D or 3D convolutional neural networks (CNNs). The mechanism of different quantization methods is studied as well. All activities are finally classified by late fusing the predictions from each clue.

\textbf{Dense-Captioning Events in Videos (ActivityNet Captions)}: For this task, we firstly adopt a standard ``detection by classification" framework to localize temporal proposals of interest in video, and then generate the descriptions for each proposal. Specifically, a two-layer LSTM-based captioning architecture with temporal attention mechanism is leveraged to generate sentence conditioning on the input video representation and its detected attributes. Moreover, the captioning architecture is equipped with policy gradient optimization scheme to further boost video captioning.

\textbf{Spatio-temporal Action Localization (AVA)}: We present a new Long Short-Term Relation Networks (LSTR), which models both short-term and long-term human-context relation to augment features for spatio-temporal action localization. Technically, Region Proposal Network (RPN) is employed to first generate bounding box proposals on the keyframe of each video clip. LSTR then models short-term human-context interactions within each clip through spatio-temporal attention mechanism and reasons long-term temporal dynamics across video clips via Graph Convolutional Networks (GCN) in a cascaded manner. The upgraded relation-aware feature of each proposal is finally employed for classifying actions.

\end{abstract}

\begin{figure*}[!tb]
\centering {\includegraphics[width=0.9\textwidth]{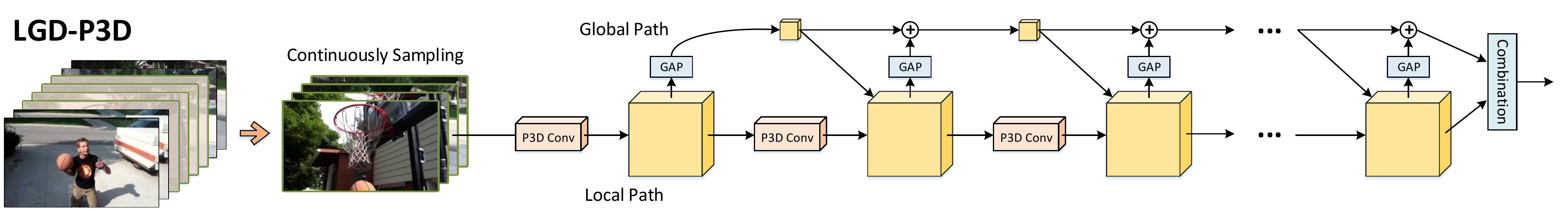}}
\caption{Network architecture of LGD-P3D. The LGD framework is proposed in \cite{qiu2019learning} and the basic P3D operation is proposed in \cite{qiu2017learning}.}
\label{fig:lgd-p3d}
\end{figure*}

\section{Introduction}
Recognizing activities in videos is a challenging task as video is an information-intensive media with complex variations. In particular, an activity may be represented by different clues including frame, short video clip, motion (optical flow) and long video clip. In this work, we aim at investigating these multiple clues to activity classification in trimmed videos, which consist of a diverse range of human focused actions. Moreover, action detection with accurate spatio-temporal location in videos, i.e., spatio-temporal action localization, is another challenging task in video understanding and we study this task in this work. Compared to temporal action localization which temporally localizes actions, this task is more difficult due to the complex variations and large spatio-temporal search space. In addition to the above two tasks tailored to activity which is usually the name of action/event in videos, the task of dense-captioning events in videos is explored here which goes beyond activities by describing numerous events within untrimmed videos with multiple natural sentences.

The remaining sections are organized as follows. Section 2 presents all the features which will be adopted in our systems, while Section 3 details the feature quantization strategies. Then the descriptions and empirical evaluations of our systems for three tasks are provided in Section 4-6 respectively, followed by the conclusions in Section 7.

\section{Video Representations}
We extract the video representations from multiple clues including frame, motion and audio.

\begin{figure}[!tb]
   \centering
   \subfigure[P3D-A]{
     \label{fig:fig1:a}
     \includegraphics[width=0.15\textwidth]{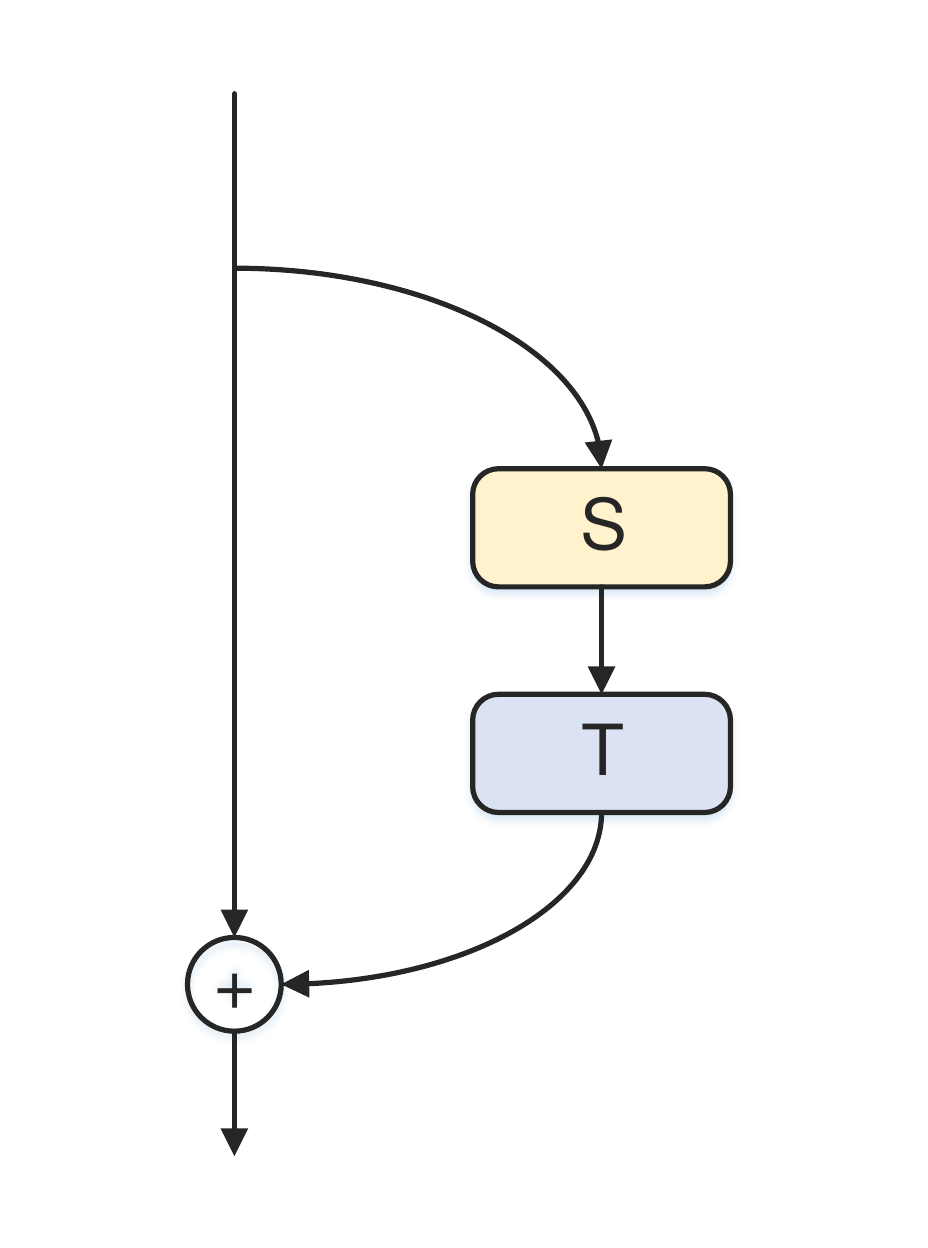}}
   \subfigure[P3D-B]{
     \label{fig:fig1:b}
     \includegraphics[width=0.15\textwidth]{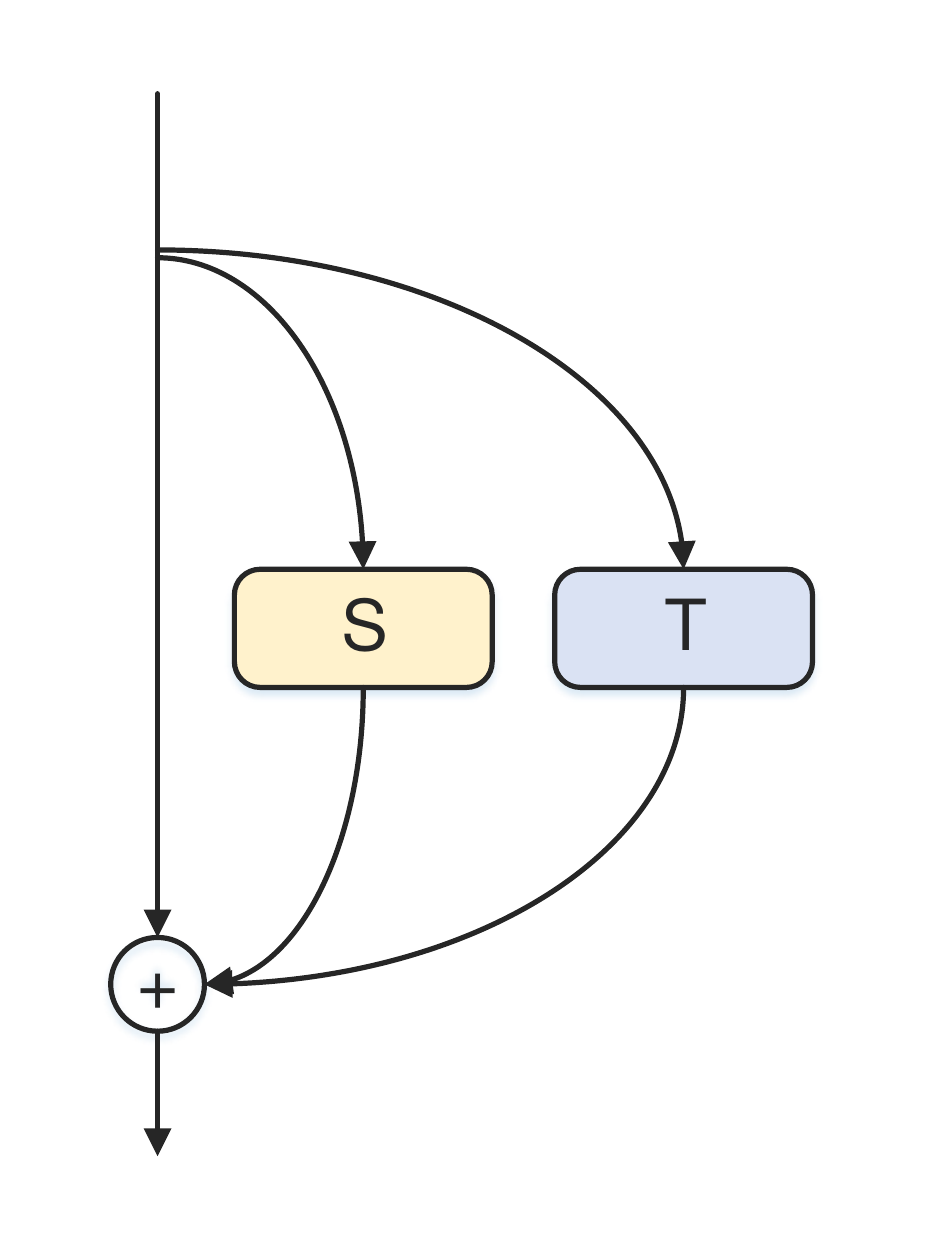}}
   \subfigure[P3D-C]{
     \label{fig:fig1:c}
     \includegraphics[width=0.15\textwidth]{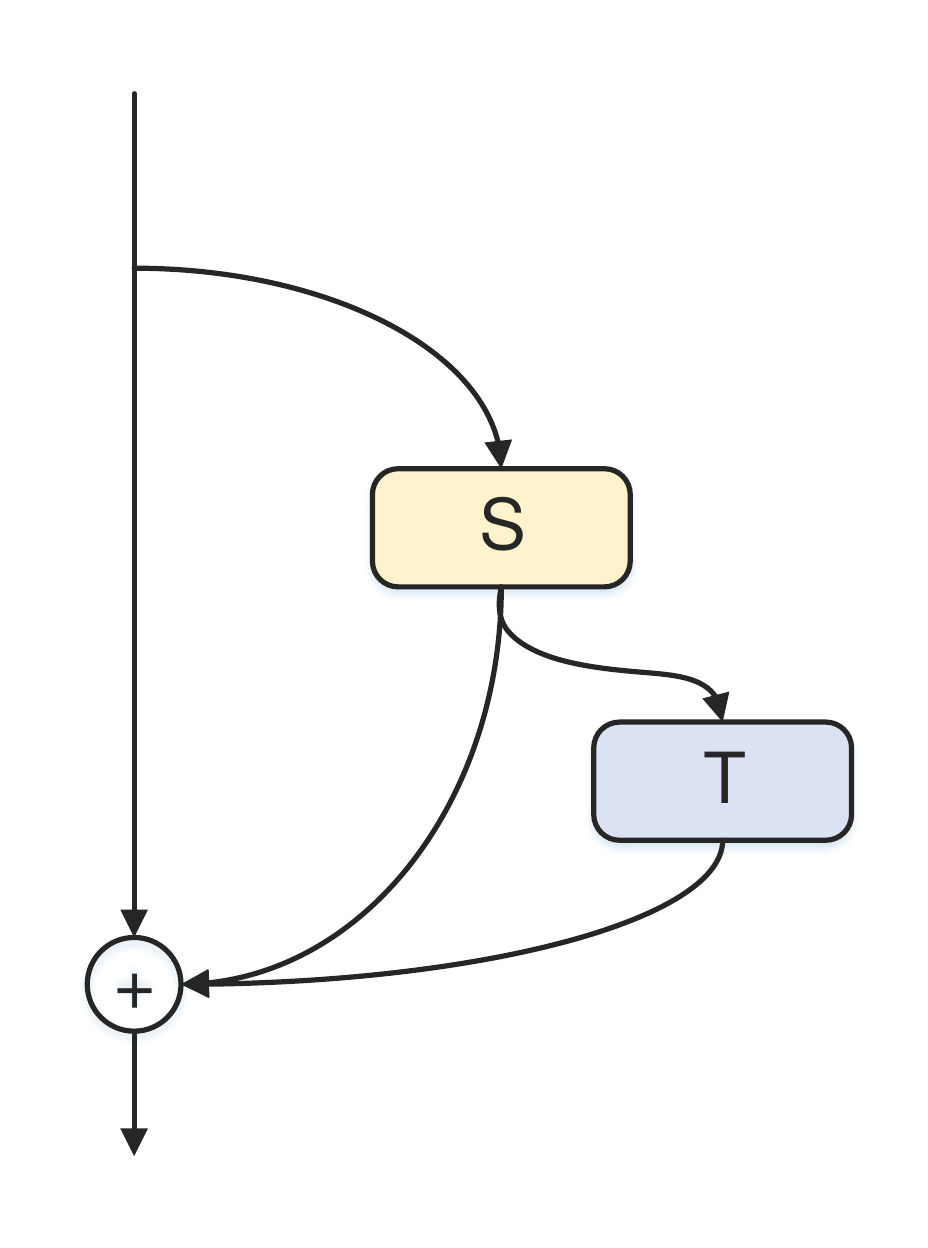}}
   \caption{\small Three Pseudo-3D blocks.}
   \label{fig:figP3D}
\end{figure}

\textbf{Frame.}
To extract appearance-based representations from video, we devise the novel Pseudo-3D Residual Net \cite{qiu2017learning} with Local and Global Diffusion \cite{qiu2019learning} (LGD-P3D) architecture, as shown in Figure \ref{fig:lgd-p3d}. The Local and Global Diffusion (LGD) is a novel neural network architecture that learns the local and global representations in parallel. The architecture is composed of LGD blocks, where each block updates local and global features by modeling the diffusions between these two representations. Diffusions effectively interact two aspects of information, i.e., localized and holistic, for more powerful way of representation learning. The basic operations in LGD-P3D are variants of bottleneck building blocks to combine 2D spatial and 1D temporal convolutions, as shown in Figure \ref{fig:figP3D}. The backbone of LGD-P3D is either ResNet-101 \cite{he2015deep} or Xception \cite{Chollet2017CVPR}. We sample 16 consecutive frames as a short clip and fix the sample rate as 2 clips per second.

\textbf{Motion.}
To model the change of consecutive frames, we apply another CNNs to optical flow ``image,'' which can extract motion features between consecutive frames. When extracting motion features, we follow the setting of \cite{qiu2019learning}, which fed a 16-frame optical flow image sequence, consisting of two-direction optical flow from multiple consecutive frames, into LGD-P3D network in each iteration. The sample rate is also set to 2 clips per second.

\textbf{Audio.}
Audio feature is the most global feature (though entire video) in our system. Although audio feature itself can not get very good result for action recognition, but it can be seen as powerful additional feature, since some specific actions are highly related to audio information. Here we utilize Xception network to extract audio feature from the audio spectrum map.

\section{Feature Quantization}
In this section, we describe two quantization methods to generate video-level representations from the extracted features.

\textbf{Average Pooling (AP).}
Average pooling is the most common method to extract video-level features. For a set of clip-level features $F=\{f_1, f_2, ..., f_N\}$, the video-level representations are produced by simply averaging all the features in the set:
\begin{equation}\label{Eq:Eq1}
\begin{array}{l}
R_{AP}=\frac{1}{N}\sum\limits_{i:f_i \in F}{f_i}
\end{array},
\end{equation}
where $R_{AP}$ denotes the final representations.

\textbf{Temporal Convolutional Pooling (TCP).}
Moreover, we utilize a novel temporal convolutional pooling to produce highly discriminative video-level representation by modeling the feature sequence with stacked 1D temporal convolutions. The video-level representations are given by:
\begin{equation}\label{Eq:Eq2}
\begin{array}{l}
R_{TCP}={Conv1D}(\{f_1, f_2, ..., f_N\})
\end{array},
\end{equation}

Here we devise a novel Conv1D network with 5 stacked depth-wise residual blocks for TCP.

\begin{figure*}[!tb]
\centering {\includegraphics[width=0.9\textwidth]{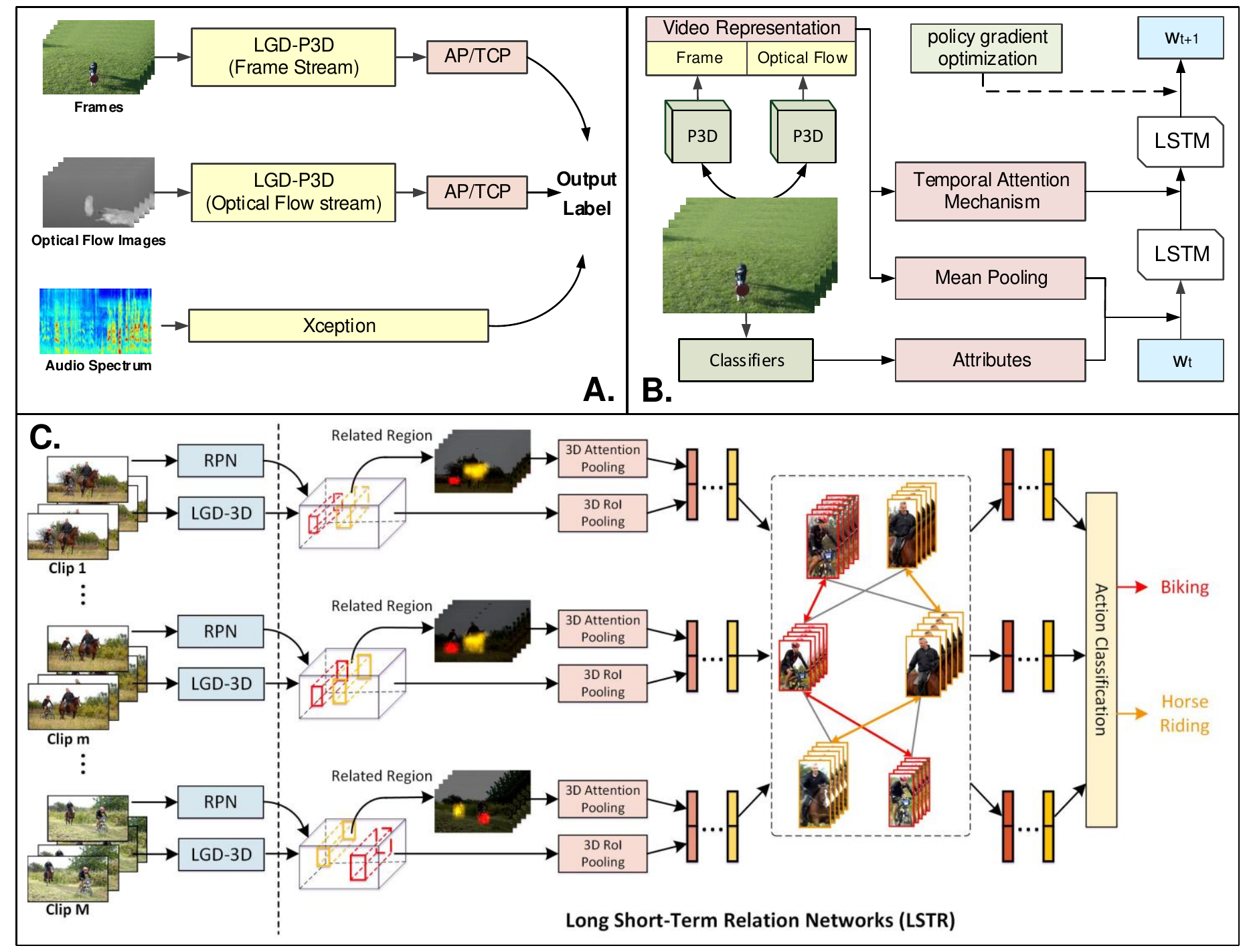}}
\caption{Frameworks of our proposed (a) trimmed action recognition system, (b) dense-captioning events in videos system, and (c) spatio-temporal action localization system.}
\label{fig:figall}
\end{figure*}

\section{Trimmed Action Recognition}
\subsection{System}
Our trimmed action recognition framework is shown in Figure \ref{fig:figall} (a). In general, the trimmed action recognition process is composed of three stages, i.e., multi-stream feature extraction, feature quantization and prediction generation. For deep feature extraction, we follow the multi-stream approaches in \cite{li2016action,qiumsr16,qiumsr,qiu2017deep}, which represented input video by a hierarchical structure including short clip, optical flow images. In addition to visual features, the most commonly used audio spectrum is exploited to further enrich the video representations. After extraction of raw features, different quantization and pooling methods are utilized on different features to produce global representations of each trimmed video. Finally, the predictions from different streams are linearly fused by the weights tuned on validation set.

\subsection{Experiment Results}
Table \ref{table:tab1} shows the performances of all the components in our trimmed action recognition system. Overall, the TCP on LGD-P3D (Xception 16-frame) achieves the highest top1 accuracy (70.63\%) and top5 accuracy (89.32\%) of single component. And by additionally apply this model on both frame and optical flow, the two-stream LGD-P3D (Xception, 16-frame\&flow) achieves an obvious improvement, which gets top1 accuracy of 72.82\% and top5 accuracy of 90.73\%. For the final submission, we linearly fuse all the components.

\begin{table*}
\centering
\caption{Comparison of different components in our trimmed action recognition framework on Kinetics validation set for trimmed action recognition task.}
\label{table:tab1}
\begin{tabular}{l|c|c|c|c} \hline
~~~\textbf{Stream}&~~~\textbf{Feature}~~~&~~~\textbf{Quantization}~~~&~~~\textbf{Top1}~~~&~~~\textbf{Top5}~~~\\ \hline
~~~\multirow{2}{*}{Frame}
& LGD-P3D (Xception, 16-frame)  & AP & 67.51\% & 86.93\% \\
& LGD-P3D (Xception, 16-frame)  & TCP & 70.63\% & 89.32\% \\
& LGD-P3D (Xception, 128-frame)  & AP & 69.84\% & 88.33\%\\
& LGD-P3D (ResNet-101, 128-frame)  & AP & 69.75\% & 88.80\%\\
\hline
~~~\multirow{3}{*}{Motion}
& LGD-P3D (Xception, 16-flow)  & AP & 54.40\% & 77.28\% \\
& LGD-P3D (Xception, 16-flow)  & TCP & 60.51\% & 82.4\% \\
& LGD-P3D (Xception, 128-flow)  & AP & 61.33\% & 83.12\%\\
& LGD-P3D (ResNet-101, 128-flow)  & AP & 64.49\% & 85.50\%\\
\hline
~~~\multirow{1}{*}{Audio}
& Xception  & AP & 21.91\% & 36.86\% \\
& Xception  & TCP & 21.76\% & 36.93\% \\
\hline
~~~\multirow{1}{*}{Two-stream}
& LGD-P3D (Xception, 16-frame\&flow)  & AP & 69.39\% & 88.08\% \\
& LGD-P3D (Xception, 16-frame\&flow)  & TCP & 72.82\% & 90.73\% \\
& LGD-P3D (Xception, 128-frame\&flow)  & AP & 71.77\% & 89.75\% \\
& LGD-P3D (ResNet-101, 128-frame\&flow)  & AP & 72.32\% & 90.46\% \\

\hline
~~~\multirow{1}{*}{Two-stream+Audio}
& LGD-P3D (Xception, 16-frame\&flow)  & AP & 70.94\% & 88.81\% \\
& LGD-P3D (Xception, 16-frame\&flow)  & TCP & 74.82\% & 91.78\% \\
& LGD-P3D (Xception, 128-frame\&flow)  & AP & 73.90\% & 90.91\% \\
& LGD-P3D (ResNet-101, 128-frame\&flow)  & AP & 74.19\% & 91.41\% \\
\hline
~~~\multirow{1}{*}{Ensemble}
&  &  & 76.37\% & 92.78\% \\

\hline
\end{tabular}
\end{table*}

\section{Dense-Captioning Events in Videos}
\subsection{System}
The main goal of dense-captioning events in videos is jointly localizing temporal proposals of interest in videos and then generating the descriptions for each proposal/video clip. Hence we firstly leverage a standard ``detection by classification" in \cite{yao2017msr} to localize temporal proposals of events in videos (5 proposals for each video). Then, given each temporal proposal (i.e., video segment describing one event), our dense-captioning system capitalizes on a two-layer LSTM-based captioning architecture with temporal attention mechanism for sentence generation. Specifically, the generative module with LSTM is inspired from the recent successes of probabilistic sequence models leveraged in vision and language tasks (e.g., image captioning \cite{li2019pointing,vinyals2015show,yao2017novel,yao2018exploring,yao2017boosting}, video captioning \cite{pan2016jointly,pan2017seeing,pan2017video}, video generation from captions \cite{pan2017to} and dense video captioning \cite{li2018jointly,yao2017msr}). We mainly utilize the two-layer LSTM-based captioning architecture in \cite{anderson2017bottom} and extend the original spatial attention at region level into temporal attention at frame level. To be specific, the first-layer LSTM collects the maximum contextual information by concatenating each input word with the previous output of second-layer LSTM, the mean-pooled video representation, and attribute representation. Next, conditioning on the output hidden state of the first-layer LSTM, a normalized temporal attention distribution over all frames is measured to dynamically fuse all frame features into attended video-level representation, which will be set as the input of the second-layer LSTM. Note that we employ the policy gradient optimization method with reinforcement learning \cite{Rennie:2016SCST} to further boost the video captioning performances specific to METEOR metric. The overall architecture of our dense-captioning system is shown in Figure \ref{fig:figall} (b).

\begin{table}[!tb]
\centering
\caption{Performance on ActivityNet captions validation and testing set. All values are reported over METEOR metric (\%).}
\label{table:tabdense}
\begin{tabular}{l|c|c}\hline
~~~\textbf{Model}&~~~~\textbf{Val}~~~~&~~~~\textbf{Test}~~~~\\ \hline
~~~\textbf{{Ours}}                                & 9.81    & -     \\\hline
~~~\textbf{{Ours + policy gradient}}   & 10.30    & 8.49 \\\hline

\end{tabular}
\end{table}

\subsection{Experiment Results}
Table \ref{table:tabdense} shows the performances of our proposed dense-captioning events in videos system. In particular, by additionally incorporating the policy gradient optimization scheme into our system, we can clearly observe the performance boost in METEOR.

\section{Spatio-temporal Action Localization}

\subsection{System}
Figure \ref{fig:figall} (c) shows the framework of spatio-temporal action localization, which includes two main components:

\textbf{Person Detector.}
We use Faster R-CNN \cite{ren2015faster} with a Deformable ResNet-101 \cite{dai2017deformable} backbone for person detection. The model is pre-trained on ImageNet \cite{karpathy2014large} and COCO \cite{lin2014microsoft}, and then fine-tuned on AVA bounding boxes. The final model obtains 93.5 AP@50 on the AVA validation set.

\textbf{Long Short-Term Relation Networks (LSTR).}
LSTR takes 8 consecutive 16-frame clips as input and employs LGD-3D ResNet-101 \cite{qiu2019learning} as backbone, which is initialized with Kinetics-600 \cite{kay2017kinetics} pre-trained model. We feed each clip to the backbone and extract the clip feature representation at the last convolutional layer. For each actor proposal, we crop and resize the clip feature within the proposal using 3D RoI Pooling to obtain a fixed-length actor representation. However, this representation ignores the short-term relation within clip representing the interactions between actors and their surroundings (including other actors, objects, and scenes). We devise a spatio-temporal attention module to model and incorporate such information into proposal representation, as illustrated in Figure \ref{fig:figall} (d). We exploit adaptive convolution to dynamically predict the actor-specific spatio-temporal attention map, which indicates the relevance degree of the global context to this actor. The context feature is then generated through 3D Attention Pooling on the attention map. The final proposal representation is obtained by concatenating the actor feature and context feature together. In addition to the short-term relation between actors and context within each clip, we also expect to further capitalize on long-range dependencies between correlated proposals from neighboring clips. To achieve this, we build a relation graph with undirected edges on human proposals extracted from all video clips. The vertex represents each human proposal and the edge denotes the relation measured on both visual similarity and geometrical overlap in between. Graph Convolutional Networks (GCN) \cite{kipf2016semi} are utilized to enrich the feature of human proposal by propagating the relation in the graph. The upgraded relation-aware feature of each proposal is finally exploited for action classification.

\subsection{Experiment Results}
Following \cite{li2018recurrent,li2018unified,long2019gaussian,qiu2017learning}, we also exploit a two-stream pipeline for utilizing multiple modalities, where the RGB frame and the stacked optical flow image are considered. To fuse the detection results, late fusion scheme is taken to average the classification scores. Table \ref{table:AVA} shows the performances of all the components in our LSTR. For the final submission, all the components are linearly fused using the weights tuned on validation set. The final mAP on validation set is 30.5\%.

\begin{table}[]
\centering
\caption{Comparison of different components in our LSTR on AVA validation/test set for spatio-temporal action localization task.}
\label{table:AVA}
\begin{tabular}{l|c|c}
\hline
\textbf{Stream}          & \textbf{Val} & \textbf{Test} \\ \hline
RGB                      & 28.3         & 27.3                \\ \hline
Flow                     & 22.7         & -                \\ \hline
Two Stream               & 29.4         & -                \\ \hline
Two Stream (multi-scale) & 30.5         & 29.1                \\ \hline
\end{tabular}
\end{table}

\section{Conclusion}
In ActivityNet Challenge 2019, we mainly focused on multiple visual features, different strategies of feature quantization and video captioning from different dimensions. Our future works include more in-depth studies of how fusion weights of different clues could be determined to boost the action recognition and spatio-temporal action localization performance. For dense-captioning events in videos task, we are targeting at making use of non-autoregressive encodeing/decoding \cite{chen2019temporal,vaswani2017attention} for sentence generation.

{
\bibliographystyle{ieee}
\bibliography{egbib}
}

\end{document}